\documentclass[letterpaper, 10 pt, conference]{ieeeconf}  

\IEEEoverridecommandlockouts                              

\usepackage{graphics,graphicx} 
\usepackage{bm, amsmath} 

\usepackage{algorithm}
\usepackage{algorithmic}
\usepackage{cite}

\title{\LARGE \bf
Underactuated Hand Design Using Mechanically Realizable Manifolds*
}

\author{Tianjian Chen$^{1}$, Maximilian Haas-Heger$^{1}$, and  Matei Ciocarlie$^{1}$
\thanks{*This work was supported in part by the NASA Early Space Innovations Program through award NNX16AD13G and by the ONR Young Investigator Program through award N00014-16-1-2026.}
\thanks{$^{1}$All authors are with the department of Mechanical Engineering, Columbia University, New York, NY 10027, USA.
        {\tt\small \{tc2764, mkh2149, matei.ciocarlie\}@columbia.edu}}%
}

\begin{document}

\setlength{\abovedisplayskip}{2mm} 
\setlength{\belowdisplayskip}{2mm} 

\maketitle
\thispagestyle{empty}
\pagestyle{empty}

\begin{abstract}
Hand synergies, or joint coordination patterns, have become an
effective tool for achieving versatile robotic grasping with simple
hands or planning algorithms. Here we propose a method to determine
the hand synergies such that they can be physically
implemented in an underactuated fashion. Given a kinematic hand model and
a set of desired grasps, our algorithm optimizes
a \textit{Mechanically Realizable Manifold} designed to be achievable
by a physical underactuation mechanism, enabling the resulting hand to
achieve the desired grasps with few actuators. Furthermore, in
contrast to existing methods for determining synergies which are only
concerned with hand posture, our method explicitly optimizes the
stability of the target grasps. We implement this method in the design
of a three-finger single-actuator hand as an example, and evaluate its
effectiveness numerically and experimentally.

\end{abstract}
\section{Introduction}

Grasping synergies \cite{santello1998postural}, or the coordination of joints, are a promising way to reduce the complexity of hands without compromising versatility. The idea of grasp synergies originates in studies of the human hand, but is also used in many robotic applications. For example, synergies can be used in the planning or control algorithms for fully-actuated hands \cite{matrone2010principal,matrone2012real,ciocarlie2009hand, wimbock2011synergy, kent2014anthropomorphic}. In addition, this notion can be embedded in the mechanical design of underactuated hands, moving some of the control intelligence to the physical mechanism. Underactuated hands have gained popularity in research for the simplicity, compliance, and ability of adaptation. More importantly in the context of this paper, they are intrinsically “synergistic” since joints are coupled.

In joint space, a hand synergy represents a low-dimensional manifold along which the hand configuration can slide. For underactuated hands, we propose the term \textit{Mechanically Realizable Manifold} to refer to such a manifold that can be physically realized by a certain mechanical design. This manifold can be altered to exhibit different shapes (corresponding to different hand behaviors), by varying the kinematic parameters. 

In addition to shaping the hand before contact, a Mechanically Realizable Manifold must also create stable grasps after making contact. Many traditional synergy-based methods (with some notable exceptions~\cite{prattichizzo2013motion}), consider postural behaviors only. Here we advance the studies on synergies to also consider stability in force domain.

In this study, we aim to tackle a problem formulated as follows. We start from: (1) a hand configuration with known kinematics (e.g, number of fingers and links, shape and dimensions of fingers) but undetermined actuation parameters (e.g. tendon routes), and (2) a set of desired grasps, without accounting for underactuation. Our goal is to design the (under)actuation parameters to match the Mechanically Realizable Manifold with the desired grasps, and at the same time, keep the grasps stable.

A straightforward way to approach this problem is to extract synergies from sample grasps using Principal Component Analysis (PCA). However, due to the physical constraints, those synergies may not be mechanically realizable in an underactuated fashion. First, certain kinematics may introduce an intrinsically nonlinear Mechanically Realizable Manifold, which can not be matched with linear PCA manifolds. Second, in a physical implementation, parameters can be constrained to certain ranges, and may not be able to reach the values that PCA result requires. Furthermore, even if the synergies from PCA result can be physically implemented, there is no guarantee that this design will lead to stable grasps. 

We propose an optimization framework combing exhaustive search and convex optimization to deal with the problems above. Our main contribution in this paper can be summarized as follows: 
\begin{itemize}
\item
To the best of our knowledge, we are the first to propose a method to search for grasping synergies which are guaranteed to be mechanically realizable in cable-driven underactuated hands under the real-world constraints. 
\item
We also formulate force equilibrium into our optimization framework so that the hand can not only reach the desired pose but also functionally grasp the target objects in a stable fashion. 
\end{itemize}
In addition to the theoretical approach, we present an evaluation of
this method on a concrete design task: building a versatile yet
compact underactuated hand for an assistive free flyer robot on the International Space Station.

\section{Related Work}

A common way to apply the idea of synergies in robotic hands is to use it in planning or control algorithms for dexterous hands. Researchers presented various studies in this category, also using ``eigen-grasps" or ``eigen-postures" to refer to synergies. For example, Matrone et al. \cite{matrone2010principal} \cite{matrone2012real} proposed a PCA-based controller for a prosthetic hand using two electromyographic (EMG) channels as the low-dimensional inputs. Ciocarlie and Allen \cite{ciocarlie2009hand} discussed the applicability of low-dimensional postural subspace in the automated grasp synthesis, and proposed a planner which takes advantage of reduced dimensionality.  Wimbock et al. \cite{wimbock2011synergy} presented a synergy level impedance control for multi-finger hands. Kent et al. \cite{kent2014anthropomorphic} showed a synergistic control method for dexterous hands where the synergies are synchronized temporally. These synergy-based methods can greatly simplify the control of dexterous hands. However, these studies mostly consider postural control without accounting for grasping force equilibrium. Also, many of these studies are limited to anthropomorphic hands in which the synergies can be extracted from human data.

\begin{figure}[t!]
\centering
\includegraphics[width=65mm]{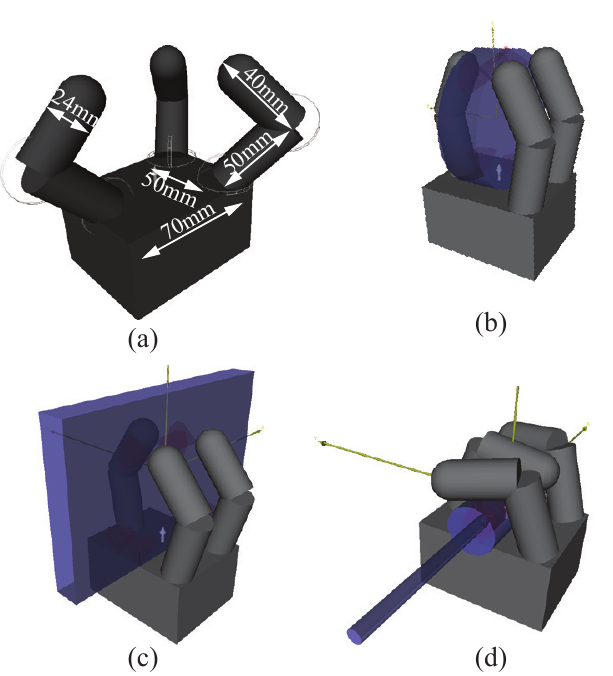}
\caption{Example of hand kinematics (a) and sample grasps (b)(c)(d)}
\label{fig:graspit}
\end{figure}

Since the methods above implement synergies in software, it is natural to think about coupling joints in hardware to reduce the number of actuators, which leads to the design of synergistic underactuated hands. For example, Brown and Asada \cite{brown2007inter} designed a mechanical implementation of PCA results for a hand using pulley-slider systems to realize inter-finger coordination. Xu et al. \cite{xu2014design} also designed a gear transmission system to enable hardware synergies for an anthropomorphic hand. These studies only consider postural behaviors without the notion of force, so they cannot guarantee grasp stability. The study from Prattichizzo et al. \cite{prattichizzo2013motion} presented analysis of the synergies with the force, but did not provide a way to determine or design the synergies. Besides, studies such as \cite{grioli2012adaptive}, \cite{catalano2014adaptive} presented the concept of ``soft synergies" and ``adaptive synergies" in underactuated hand design accounting for grasp force equilibrium, and they proposed the dexterous hand design (the Pisa/IIT hand) using these models. Though all studies above present feasible ways to implement synergies, they require additional mechanisms (e.g. pulley-sliders, gears, or differential mechanisms) dedicated exclusively to the implementation of synergies. In contrast, we consider it preferable if synergies can be realized by just optimizing some must-have mechanical components such as tendon routing points.

Optimization is a powerful tool to realize desired performance for underactuated hands. There is a lot of literature in this category, for example, the design of Harvard Hand presented by Dollar and Howe \cite{dollar2011joint} optimized the grasping range and contact forces for a planar gripper. The Velo gripper \cite{ciocarlie2014velo} was optimized to achieve both fingertip grasp and enveloping grasp using a single actuator. The iHY hand \cite{odhner2014compliant} was optimized for versatile grasps. Ciocarlie and Allen \cite{ciocarlie2011constrained} formulated the parameter design problem as a gloabal quadratic programming. Dong et. al \cite{dong2018geometric} optimizes the dimensions and tendon routes of a tendon-driven hand using Genetic Algorithm. Except for few kinematic formulations~\cite{ciocarlie2011constrained}, it is still not easy to formulate the parameter selection as a globally convex optimization with well-established algorithms, therefore researchers need to rely on brute-force search or stochastic global optimizers.

\section{Optimization Method}

\begin{figure}[t!]
\centering
\includegraphics[width=65mm]{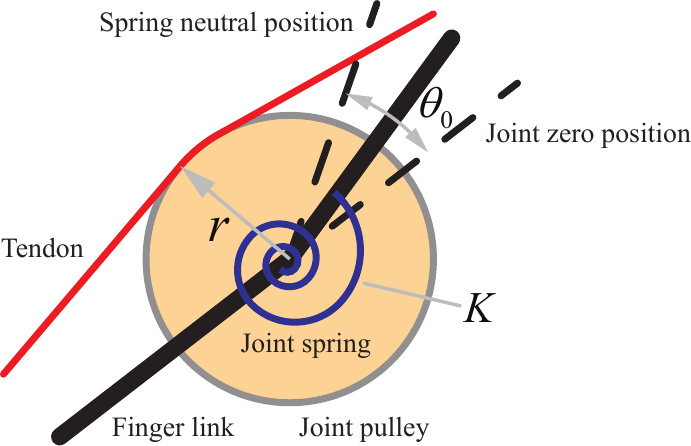}
\caption{Joint Parameters}
\label{fig:joint_params}
\end{figure}

\subsection{Problem Formulation}

Since the design space of robot hands is huge, we narrow down our problem to require a kinematic configuration beforehand, i.e. the number of fingers and links, and the shape and dimensions of fingers and the palm need to be specified. We also assume a commonly-used actuation scheme in which tendons actuate finger flexion and springs are responsible for extension. The unknowns are the actuation parameters.

We also assume we are given a set of desired grasps (e.g. Fig.\ref{fig:graspit}). These grasps do not account for underactuation, i.e. the joints are considered independent. Furthermore, we require that all these sample grasps have force-closure property, i.e. a set of contact forces exist that produce a zero resulting wrench on the object while satisfying friction constraints. 

Our goals are as follows. We are aiming to optimize the (under)actuation
parameters under physical constraints to reshape the corresponding Mechanically Realizable Manifold (on which the hand configuration point can slide) to reach the desired grasps. Furthermore, we
need to maximize the stability of the desired grasps, accounting for
underactuation, after the contacts are made.

The actuation parameters we wish to optimize are: the tendon moment arms (or pulley radii) $r$ around joints, the spring stiffnesses $K$, and the spring preload angles $\theta_0$ (which are defined as the spring angles when joint angles are zeros), shown in Fig. \ref{fig:joint_params}. Considering the number of joints in a versatile hand, this can be a high dimensional optimization.

In addition, the optimization needs to obey certain constraints, and the ability to deal with physical constraints is one of the key advantages of our method compared to simple application of PCA. For example, the restoring spring stiffnesses are limited by the physical dimensions of mounting space allowed in the finger designs, and are only available in a discrete series of values offered by the vendor, and the joint restoring torques should be within the range that the actuators can overcome, etc.

\subsection{Problem Decomposition}

We have two goals in our problem: optimizing the kinematic behaviors before contact, and force-generation behaviors after contact. In theory, both kinematic behaviors and force equilibrium are related to all actuation parameters listed above. However, this will result in two objective functions in the same design space. While formulating some (weighted) combination of these objectives is possible, it would require arbitrarily assigned weights, which we would like to avoid; the joint optimization is also high-dimensional. It would thus be beneficial if we could split the kinematic optimization and force optimization in an appropriate way. 

We found that after the hand makes initial contact with an object, the net joint torques, which determine grasp equilibrium, are only related to tendon moment arms. For a single joint, equilibrium when contacts are just made can be expressed as
\begin{equation} \label{eq:eq_before_touch}
r(\theta)t_b - K(\theta+\theta_{0}) = 0
\end{equation}
where $r$, $t_b$ are the tendon moment arm and tension (subscript $b$ means "before contact"), $K$ is the joint spring stiffness, and $\theta$ and $\theta_0$ are the joint angle and spring preload angle. Once additional torque is applied and the grasp is loaded, equilibrium can be expressed as 
\begin{equation} \label{eq:eq_after_touch}
r(\theta)t_a - K(\theta+\theta_0) = \tau_{net} 
\end{equation}
where $t_a$ is the tendon tension, and subscript $a$ means "after applying torque". Combining Eqs. (\ref{eq:eq_before_touch}) and (\ref{eq:eq_after_touch}) suggests that the net joint torque is only affected by the moment arms:
\begin{equation} \label{eq:net_trq}
\tau_{net} = r(\theta)(t_a-t_b) = r(\theta) t_{net}
\end{equation}

This means we can decompose the problem as follows. First, we assume these grasps can be achieved from a hand posture perspective, and then optimize the tendon moment arms for the equilibrium of objects and hand in the grasp. After that, we optimize the rest of the actuation parameters (spring stiffnesses and preload angles) to make sure that the sample grasps can actually be reached as closely as possible.

%
%

\subsection{Force Optimization}
The goal of the force optimization is to have the underactuated hand and object as close to equilibrium as possible in the grasping phase. As discussed before, the variables we optimize are the \textit{tendon moment arms} in every joints.

\subsubsection{Grasp Analysis}

To form an objective for the optimization, an evaluation of grasp stability is needed. Here we employ a commonly-used grasp analysis formulation \cite{ciocarlie2011constrained}, \cite{miller2003implementation}. We use the (linearized) model of point-contact with friction.  For contact $k$, contact wrench $\bm{c}_k$ can be expressed as linear combinations of $\bm{\beta}_k$ --- the amplitudes of the frictional and normal components, related by the matrix $\bm{D}_k$, as shown in (\ref{eq:contact_wrench}). Equations (\ref{eq:local_friction_constraint}) and (\ref{eq:beta_k_bound}) model the effect that contact force must be constrained inside the friction cone (or pyramid). The details about the construction of matrices $\bm{D}_k$ and $\bm{F}_k$ can be found in \cite{prattichizzo2008grasping}.

\begin{equation} \label{eq:contact_wrench}
\bm{c}_k = \bm{D}_k \bm{\beta}_k
\end{equation}

\vspace{-2mm}

\begin{equation} \label{eq:local_friction_constraint}
\bm{F}_k \bm{\beta}_k \leq \bm{0}
\end{equation}

\vspace{-2mm}

\begin{equation} \label{eq:beta_k_bound}
\bm{\beta}_k \geq \bm{0}
\end{equation}

In general, a grasp is stable if the following conditions are satisfied:
\begin{itemize}
\item
\textit{Hand equilibrium}: the active joint torques are balanced by contact forces. 
\begin{equation} \label{eq:hand_eq}
\bm{J}^T\bm{c} = \bm{J}^T\bm{D\beta} = \bm{\tau}_{eq}
\end{equation}

\item
\textit{Object equilibrium}: the resultant object wrench is zero.
\begin{equation} \label{eq:obj_eq}
\bm{Gc} = \bm{GD\beta} = \bm{0}
\end{equation}

\item
\textit{Friction constraints}: the contact forces are constrained inside the friction cone.
\begin{equation} \label{eq:friction_constraint}
\bm{F \beta} \leq \bm{0}
\end{equation}

\vspace{-2mm}

\begin{equation} \label{eq:beta_bounds}
\bm{\beta} \geq \bm{0}
\end{equation}

\end{itemize}

In these formulations, $\bm{J}$ is the contact Jacobian, $\bm{G}$ is the grasp map matrix, $\bm{D}$ and $\bm{F}$ are block-diagonal matrices constructed by $\bm{D}_k$ and $\bm{F}_k$ respectively, $\bm{c}$ and $\bm{\beta}$ are stacked vectors constructed by $\bm{c}_k$ and $\bm{\beta}_k$ respectively, and $\bm{\tau}_{eq}$ is the joint torque vector to create hand equilibrium.

\subsubsection{Force Optimization Formulation}

In our problem, since the hand is underactuated, the entries in joint torque vector are not independent. Instead, $\bm{\tau}_{real}$ follows the relationship:
\begin{equation} \label{eq:actuation}
\bm{\tau}_{real} = \bm{A}(r_1, r_2, \cdots)\bm{t}_{net}
\end{equation}
where $\bm{A}$ is the actuation matrix, which is a function of the tendon moment arms. $\bm{t}_{net}$ is the net tendon tension vector, in which each element represents the incremental tension of an independent tendon (whose force is not related to others) comparing to the tension just before touch.

For a set of given tendon moment arms and a given grasp pose, we wish to find the contact force magnitudes $\bm{\beta}$ and the net tension in each independent tendon $\bm{t}_{net}$, which solve (\ref{eq:hand_eq})--(\ref{eq:actuation}). We change the search for the exact solution to an optimization problem by turning hand equilibrium (\ref{eq:hand_eq}) from a constraint into an objective function. We use the norm of the unbalanced joint torques $\Delta\bm{\tau}_{a}$(shown in (\ref{eq:unbalanced_trq_grasping}), subscript $a$ meaning "after contact") as the stability metric, and lower value is considered better.
\begin{equation} \label{eq:unbalanced_trq_grasping}
\Delta\bm{\tau}_{a} = \bm{\tau}_{eq} - \bm{\tau}_{real} = \bm{J}^T\bm{D\beta} - \bm{At}_{net}
\end{equation}
The problem to find the minimal norm of unbalanced torques is a convex Quadratic Program (QP), shown in Algorithm \ref{alg:qp1}.

\begin{algorithm}[t]
\caption{Calculate grasping phase stability metric (QP)}
\label{alg:qp1}
\begin{algorithmic}[]


\STATE{find: $ \bm{x} = \begin{bmatrix}\bm{\beta} \\ \bm{t}_{net}\end{bmatrix} $}
\vspace{2mm}

\STATE{minimize: $ \lVert \Delta\bm{\tau}_{a} \rVert^2 = \lVert \bm{Qx} \rVert^2 = \bm{x}^T\bm{Q}^T\bm{Qx}$
\vspace{2mm}

where $ \bm{Q} = \begin{bmatrix}\bm{J}^T\bm{D} & \bm{-A}\end{bmatrix} $}
\vspace{2mm}

\STATE{subject to:}

\vspace{-2mm}

\begin{equation} \label{eq:obj_eq_extended}
\begin{bmatrix}\bm{GD} & \bm{O}\end{bmatrix}\bm{x} = \bm{0}
\end{equation}

\vspace{-2mm}

\begin{equation} \label{eq:friction_cons_extended}
\begin{bmatrix}\bm{F} & \bm{O}\end{bmatrix}\bm{x} \leq \bm{0}
\end{equation}

\vspace{-2mm}

\begin{equation} \label{eq:x_bound}
\bm{x} \geq \bm{0}
\end{equation}

\vspace{-2mm}

\begin{equation} \label{eq:sum_trq}
\begin{bmatrix}1 \cdots 1\end{bmatrix}\begin{bmatrix}\bm{J}^T\bm{D} & \bm{O}\end{bmatrix}\bm{x} = 1
\end{equation}

\end{algorithmic}
\end{algorithm}
The constraints(\ref{eq:obj_eq_extended})--(\ref{eq:x_bound}) are extended versions of (\ref{eq:obj_eq})--(\ref{eq:beta_bounds}) and the last one (\ref{eq:sum_trq}) prevents the trivial solution where all contact forces and joint torques are zeros, by constraining the sum of joint torques to be a non-zero number.

We formulated the search for optimal tendon moment arms as a dual-layer optimization. The outer-layer is the exhaustive search over all combinations of the discretized tendon moment arms, and the inner layer calculates the stability metric using the QP in Algorithm \ref{alg:qp1} over all sample grasps. Though the inner-layer is convex, the outer-layer is not, and is not trivial to be reformulated as a convex problem. Therefore we choose to discretize the unkowns and search by brute force, as illustrated in Algorithm \ref{alg:force_optim}.

\begin{algorithm}[t]
\caption{Force Optimization}
\label{alg:force_optim}
\begin{algorithmic}[1]
	\FOR{each combination of the tendon moment arms ($r$)}
		\FOR{each sample grasp $i$}
			\STATE{Calculate the stability metric $q_i = \lVert \Delta\bm{\tau}_{a} \rVert$ in grasping phase using Algorithm \ref{alg:qp1}}
		\ENDFOR
	\STATE {Calculate the overall stability metric $Q = \lVert \left[q_1, q_2, \cdots \right] \rVert$}	
	\ENDFOR
	\STATE {Select the combination of the tendon moment arms with minimum $Q$}

\end{algorithmic}
\end{algorithm}

\subsection{Kinematic Optimization}
The goal of the kinematic optimization is to make sure the Mechanically Realizable Manifolds are as close to the sample grasps as possible. We use the tendon moment arms optimized in the previous subsection. Therefore the dimension of the design space is reduced, and the remaining parameters to be optimized are: the spring stiffnesses ($K$) and spring preload angles ($\theta_0$) for all joints.

We translate the goal of reaching sample grasps quasi-statically to the one of maximizing stability of the sample grasp configurations. Higher sample grasp stability means the grasps are closer to Mechanically Realizable Manifold, on which the hand is in quasi-static equilibrium. We use the unbalanced joint torques before contact in the sample grasp poses as the stability metric. We emphasize that in this part we only consider the equilibrium of the hand itself, without the object. We also note that the pool of sample grasps are different from the previous section: in addition to the sample poses, we also include the fully open configuration and give it high weights, to make sure that the hand can actually open, instead of moving between sample configurations.

Given certain parameters and certain poses, the unbalanced torque of a certain joint can be written the same way as (\ref{eq:eq_after_touch}), but $\bm{\tau}_{net}$ is now unbalanced. For multiple joints connected by the same tendon(s), the unbalanced torque vector $\Delta\bm{\tau}_{b}$ (subscript $b$ means "before contact") can be calculated as:
\begin{equation} \label{eq:unbalanced_trq_free_motion}
\Delta\bm{\tau}_{b}  = \bm{R}(r_1, r_2, \cdots)\bm{t} - \bm{\tau}_s
\end{equation}
where the matrix $\bm{R}$ is a kinematic-dependent known function of $r_1, r_2, \cdots$ (the tendon moment arms optimized in the previous subsection), $\bm{\tau}_{s} = [K_1(\theta_1 + \theta_{01}), K_2(\theta_2 + \theta_{02}), \cdots]^T $ is a known vector of spring torques calculated by the given spring parameters and given pose, and $\bm{t}$ is a unknown vector of the tensions. We wish to find the $\bm{t}$ vector which results in the minimum norm of unbalanced joint torques, which is also a convex QP as shown in Algorithm \ref{alg:qp2}.

\begin{algorithm}[t]
\caption{Calculate stability metric before contact (QP)}
\label{alg:qp2}
\begin{algorithmic}[]

\vspace{2mm}
\STATE{find: $\bm{t}$}

\vspace{2mm}
\STATE{minimize: $\lVert\Delta\bm{\tau}_{b}\rVert^2 = \bm{t}^T\bm{R}^T\bm{Rt} - 2\bm{\tau}_{s}^T\bm{Rt} + \bm{\tau}_{s}^2 $}

\vspace{2mm}
\STATE{subject to: $\bm{t} \geq \bm{0}$}

\vspace{2mm}

\end{algorithmic}
\end{algorithm}

\begin{algorithm}[t]
\caption{Kinematic Optimization}
\label{alg:kinematic_optim}
\begin{algorithmic}[1]
	\FOR{each combination of the spring stiffnesses ($K$) and preload angles ($\theta_0$)}
		\FOR{each sample grasp $i$}
			\STATE{Calculate the stability metric  $q_i = \lVert \Delta\bm{\tau}_{b} \rVert$ before contact using Algorithm \ref{alg:qp2}}
		\ENDFOR
	\STATE {Calculate the overall stability metric $Q = \lVert \left[q_1, q_2, \cdots \right] \rVert$} 		
	\ENDFOR
	\STATE {Select the optimal combination of $K$'s and $\theta_0$'s}

\end{algorithmic}
\end{algorithm}

To search the optimal spring stiffnesses and preload angles, we also formulated a dual-layer framework, where the outer-layer is an exhaustive search over all combinations of spring stiffnesses and preload angles, and the inner-layer is to calculate the stability metric (unbalanced joint torques before contact) for all grasps. Similar to the Force Optimization, the outer-layer is also a non-convex problem. The pseudo code is shown as Algorithm \ref{alg:kinematic_optim}.

One side note for both force and kinematic optimization is that, in practice, the exhaustive search in the outer loop may have redundant degrees-of-freedom, because the global optimum may not be unique, but may instead form a manifold, depending on the given kinematic form (for example, if only the ratio of certain parameters matter as opposed to their absolute values). If this is the case, we have the freedom to arbitrarily assign a subset of the parameters. 

\section{Optimization Example and Result}

\subsection{Design Objective}

We implemented the proposed method in the design of a 3-finger single-actuator underactuated manipulator, which is built to be used on the \textit{Astrobee} free-flying robot \cite{bualat2015astrobee} in the International Space Station (ISS).

Our method is suitable for this design task for several reasons. First, the limited room, power supply, and control inputs require a minimal number of actuators, but the tasks require versatile grasping ability; these competing goals can be achieved by highly underactuated synergistic grasping. Second, the objects in ISS are known and relatively unchanged, so including them in the sample grasps has a good chance to result in good performance in practice. Third, since the available room to store this hand in the Astrobee robot is given, the dimensions of the hand can be specified beforehand, which is what we require in our method.

\subsection{Design Process}

\subsubsection{Grasp Collection}

We selected a kinematic design which consists of three fingers, each finger consists of two links, and the thumb has two degrees-of-freedom while each of the opposing fingers has three degrees-of-freedom with a roll-pitch proximal joint. The kinematics and dimensions of the hand are shown in Fig. \ref{fig:graspit} (a). 

We modeled the hand in \textit{GraspIt!} \cite{miller2004graspit} simulator and created 21 sample grasps for 15 commonly-used objects in ISS, mainly including the food packages (such as cans) and tools (such as screw drivers). Some of the grasps are shown in Fig. \ref{fig:graspit} (b)(c)(d). All grasps have force-closure property, checked using the Ferrari-Canny $\epsilon$ metric \cite{ferrari1992planning} with $\epsilon > 0$.

\subsubsection{Force Optimization}
\begin{figure}[t!]
\centering
\includegraphics[width=70mm]{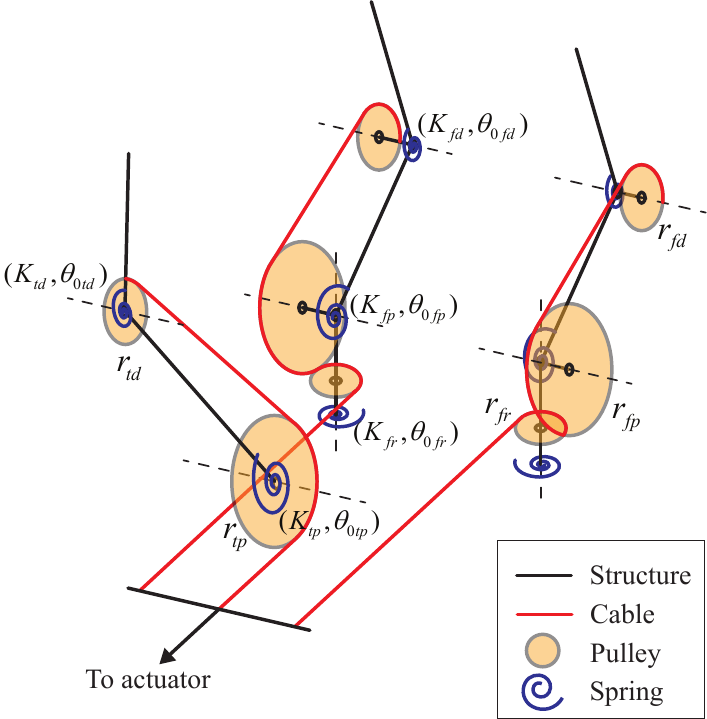}
\caption{Actuation scheme}
\label{fig:actuation}
\end{figure}

In this step, we optimize the joint pulley radii (tendon moment arms) $r_{tp}, r_{td}, r_{fr}, r_{fp}, r_{fd}$, where the subscript $t$ and $f$ represent thumb and finger, and $r$, $p$, and $d$ represent the roll, proximal and distal joints (we consider the two fingers are just mirrored versions of each other). These parameters are illustrated in Fig. \ref{fig:actuation}

The actuation scheme we designed is also shown in Fig. \ref{fig:actuation}, where each finger is actuated by one cable, and all cables are rigidly connected to the actuator. The Actuation Matrix in (\ref{eq:actuation}) has the specific form of
\begin{equation} \label{eq:actuation_matrix}
\bm{A} = \left[ \begin{smallmatrix}
r_{tp} & & \\
r_{td} & & \\
 &-r_{fr} & \\
 &r_{fp} & \\
 &r_{fd} & \\
 & &r_{fr} \\
 & &r_{fp} \\
 & &r_{fd} \\
\end{smallmatrix} \right]
\end{equation}

For each finger, only the ratio (instead of the absolute values) of the pulley radii matters. Assuming we find the optimal pulley radii, we can also vary these numbers (together with the tensions) without changing objective function value, as long as we keep the joint torques $\bm{\tau = At_{net}}$ the same (for example, scale pulley radii by a number $a$ and scale the tension by $1/a$). Therefore we have the freedom to assign one of pulley radii in each finger without loosing generality.

In the outer-layer of the algorithm \ref{alg:force_optim}, the mechanically feasible range of pulley radii is constrained to 2mm to 12mm while the step size is selected as 0.5mm. We set the proximal joint pulley radii to be 12mm according to the reasoning above. The optimized pulley radii are shown as table \ref{table:parameters}. The computation time on a commodity desktop computer (quad-core 3.40GHz CPU) is approximately 30 minutes.

\subsubsection{Kinematic Optimization}
In this step, we optimize the spring stiffnesses $K_{tp}, K_{td}, K_{fr}, K_{fp}, K_{fd}$, and the spring preload angles $\theta_{0tp}, \theta_{0td}, \theta_{0fr}, \theta_{0fp}, \theta_{0fd}$, where the subscripts have the same meaning as the ones in the previous part. These parameters are also illustrated in Fig. \ref{fig:actuation}.

Since the free motion behavior of each finger is independent from every other one, we can search for each finger separately. The matrix function $\bm{R}$ in (\ref{eq:unbalanced_trq_free_motion}) has form of $ \bm{R} = [r_{tp}, r_{td}]^T $ for the thumb and $ \bm{R} = [r_{fr}, r_{fp}, r_{fd}]^T$ for the fingers.

The spring stiffnesses are constrained mechanically by the physical dimensions to be fit in the available mounting space in each joint, and they can only take values in discrete numbers offered by the vendor. In our design, the possible range of the stiffnesses is 1.80 Nmm/rad to 6.82 Nmm/rad. Also, the spring preload angles are limited between the maximum allowed torsional angles and the minimum torsion angles to provide restoring torque over the entire range of motion. In our case, the preload angles range from $\pi/4$ to $7\pi/4$ radians for the roll joints, $\pi/4$ to $3\pi/2$ radians for the proximal joints, and $0$ to $3\pi/2$ radians for the distal joints. We divide each of the intervals evenly into 30 steps. 

Similar to the reasoning in previous part, only the stiffness ratio matters, and we have the freedom to arbitrarily assign one of the parameters for each finger. Thus we select 6.82 Nmm/rad as the proximal joint stiffnesses. The optimized stiffnesses and preload angles are shown in Table \ref{table:parameters}. The computation time including both thumb and finger optimization is approximately 15 minutes.

\begin{table}[b]
\caption{Optimized parameters (in mm, Nmm/rad, rad, respectively)} 
\label{table:parameters}
\vspace{-3mm}
\begin{center}
\begin{tabular}{c|ccccc}
Parameter&$r_{tp}$ & $r_{td}$ & $r_{fr}$ & $r_{fp}$ & $r_{fd}$\\
\hline
\\[-2.5mm]
Value&12.0 & 4.0 & 2.0 & 12.0 & 4.0\\
\end{tabular}
\end{center}

\begin{center}
\begin{tabular}{c|ccccc}
Parameter&$K_{tp}$ & $K_{td}$ & $K_{fr}$ & $K_{fp}$ & $K_{fd}$\\
\hline
\\[-2.5mm]
Value&6.82 & 1.80 & 1.80 & 6.82 & 2.11\\
\end{tabular}
\end{center}

\begin{center}
\begin{tabular}{c|cccccc}
Parameter&$\theta_{tp}$ & $\theta_{td}$ & $\theta_{fr}$ & $\theta_{fp}$ & $\theta_{fd}$\\
\hline
\\[-2.5mm]
Value&4.441 & 4.712 & 3.385 & 4.712 & 4.225\\
\end{tabular}
\end{center}
\end{table}

\begin{figure}[t!]
\centering
\includegraphics{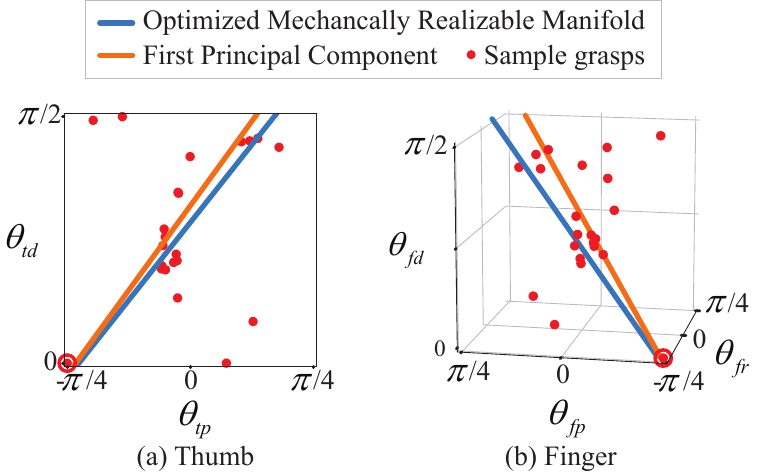}
\caption{Mechanically Realizable Manifolds vs. principal components}
\label{fig:mrm_pca}
\end{figure}

\subsection{Numerical Evaluation}
To demonstrate the effectiveness of our method, we compare the objective function values between the optimized and unoptimized cases, in which all parameters are valued in the middle of the feasible ranges described above. 
For force optimization, the norm of unbalanced torque vector over all grasps is 0.92 Nm for unoptimized case and 0.23 Nm for optimized case, meaning a 75\% reduction. For kinematic optimization, the norm of unbalanced torque vector is 96.74 Nm for unoptimized case and 7.74 Nm for optimized case, meaning a 92\% reduction.

It is also a useful practice to compare the Mechanically Realizable Manifolds with the principal component vectors from PCA. We derive the expression for the Mechanically Realizable Manifolds by setting the unbalanced joint torques in (\ref{eq:unbalanced_trq_free_motion}) to be zeros, which means that certain point in joint space can be reached quasi-statically. This results in a manifold of joint angles parameterized by cable tension. In our design case, the Mechanically Realizable Manifold of the thumb has the form of:

\begin{equation} \label{eq:thumb_manfold}
\begin{bmatrix}
\theta_{tp} \\ 
\theta_{td} 
\end{bmatrix}
= \begin{bmatrix}
\frac{r_{tp}^*}{K_{tp}^*} \\
\frac{r_{td}^*}{K_{td}^*} \\
\end{bmatrix}t_t
-\begin{bmatrix}
\theta_{0tp}^* \\
\theta_{0td}^*
\end{bmatrix}
\end{equation}

Similarly, the Mechanically Realizable Manifold of the finger has the form of:
\begin{equation} \label{eq:finger_manfold}
\begin{bmatrix}
\theta_{fr} \\ 
\theta_{fp} \\ 
\theta_{fd} 
\end{bmatrix}
= \begin{bmatrix}
\frac{r_{fr}^*}{K_{fr}^*} \\
\frac{r_{fp}^*}{K_{fp}^*} \\
\frac{r_{fd}^*}{K_{fd}^*} \\
\end{bmatrix}t_f
-\begin{bmatrix}
\theta_{0fr}^* \\
\theta_{0fp}^* \\
\theta_{0fd}^*
\end{bmatrix}\end{equation}

These manifolds are straight lines parameterized by $t_t$ or $t_f$ (the elements with superscript * are fixed values). Fig. \ref{fig:mrm_pca} shows the optimized Mechanically Realizable Manifolds together with the first principal component from PCA, in the joint spaces of the thumb and the finger. The axes represent the joint angles, the blue and orange lines represent the Mechanically Realizable Manifolds and the first principal components respectively, and the red dots are the sample grasps, and the circled red dots are the fully open positions.

\begin{figure}[t!]
\centering
\includegraphics[width=82mm]{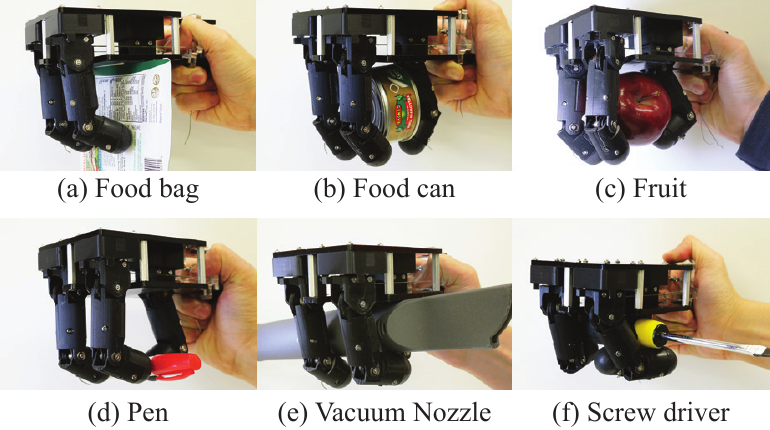}
\caption{Grasp examples using prototype hand}
\label{fig:grasp_photos}
\end{figure}

\begin{figure*}[t!]
\centering
\includegraphics[width=1.0\textwidth]{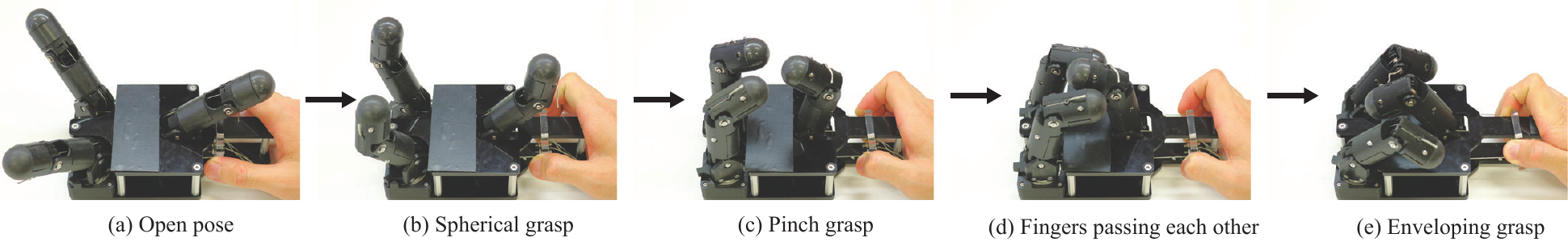}
\caption{Finger trajectory and the types of the grasp along the trajectory, also shown in the multimedia (video) attachment}
\label{fig:finger_traj}
\end{figure*}

\subsection{Construction of the Hand and Experimental Evaluation}
We constructed a physical prototype according to the parameters optimized above. Since this prototype is just intended to be proof-of-concept, it does not include a motor but is actuated by pulling the common end of the cables.

The behaviors of the hand are generally similar to the theory. Fig. \ref{fig:finger_traj} (and the accompanying multimedia attachment) demonstrates the finger trajectories, in which the hand first closes toward the center, making a spherical grasp posture and then a pinch grasp posture. Continuing to close, the fingertips do not collide but rather pass each other (due to motion in the roll degree of freedom). Finally, the hand creates an enveloping grasp shape. Fig. \ref{fig:grasp_photos} shows several grasps, displaying the versatility of the hand. We can see the hand can perform stable pinch grasps (a)(d),spherical grasps (b)(c), and power grasps (e)(f).

\section{Discussion and Conclusion}
Both the theoretical and experimental results demonstrate the effectiveness of our method: the proposed optimization framework to design underactuated hands can indeed shape the Mechanically Realizable Manifold to be as close to the desired grasps as possible, and also maximize grasp stability.


The comparison with PCA result in Fig.\ref{fig:mrm_pca} also illustrates the performance of our method. The principal components explain the most variance of the sample grasps. However, due to the mechanical constraints, this manifold cannot be reached. In contrast, the proposed method calculates a manifold which attempts to approach the PCA result as much as possible under the physical constraints. Additionally, it optimizes grasping stability. We note that here the Mechanically Realizable Manifold is one-dimensional because we selected a single-actuator design, and is linear here because we used pulleys whose radii are configuration-independent. These are not necessarily true in general.

We believe the dual-layer framework combining the non-convex global search and the convex optimization is a useful formulation. In our design case, it is not possible to formulate the problem as a global quadratic program similar to \cite{ciocarlie2011constrained}, as this results in higher-order equality constraints, which cannot be handled by currently available solvers. In contrast, the dual-layer framework has the freedom to deal with various kinematic form and actuation method. The brute force search in the outer-layer could also be replaced by stochastic global optimization algorithms.

Even though our method does not solve the problem of initial kinematic design (we require a pre-specified kinematic form), it can be used to compare different kinematic designs using the evaluation metrics from the optimization. For example, we can try designs with more actuators to tackle the limitation shown in Fig. \ref{fig:mrm_pca} (some of the sample grasps are not hit by the one-dimensional Mechanically Realizable Manifold). After testing different designs, we can compare them using the metrics in the optimization. 

Overall, we claim that the proposed framework is an effective tool to design underactuated hands moving on optimized Mechanical Realizable Manifolds and ending up with stable grasps. Future work includes the implementations of this method based on other kinematic designs (such as two-actuator design with yaw-pitch fingers), as well as the exploration on more efficient global search algorithms. We aim to further investigate these possibilities.

\addtolength{\textheight}{-12cm}   






\bibliographystyle{bib/IEEEtran}  
\bibliography{bib/sensing,bib/design,bib/control,bib/analysis,bib/planning}  

\end{document}